\documentclass{article}
\usepackage{spconf,amsmath,graphicx,booktabs,xcolor,microtype,xspace}
\usepackage{multirow}
\usepackage{balance}
\usepackage{url}
\usepackage{hyperref}
\hypersetup{hidelinks,hypertexnames=false}
\newcommand{\method}{SemReward-VL\xspace}

\title{SEMREWARD-VL: SEMANTIC REWARD-GUIDED VIDEO-LANGUAGE ADAPTATION FOR DEVELOPMENTAL BEHAVIOR ASSESSMENT}
\name{{\fontsize{9}{10}\selectfont De Jiang$^{1}$,\hspace{0.15em}Shuo Zhang$^{2}$,\hspace{0.15em}Kehong Yuan$^{1}$,\hspace{0.15em}Hongen Liao$^{2,*}$\thanks{$^{*}$Corresponding author: Hongen Liao (\nolinkurl{liao@tsinghua.edu.cn}).}}}
\address{\vspace{-0.35em}\parbox{0.95\textwidth}{\centering\fontsize{9}{9.7}\selectfont
$^{1}$Tsinghua Shenzhen International Graduate School, Tsinghua University, Shenzhen, China; 
$^{2}$School of Biomedical Engineering, Tsinghua University, Beijing, China.}}

\begin{document}
\ninept
\maketitle
\vspace{-1.0em}

\begin{abstract}
Developmental screening videos show how children perform specific behaviors, but clinical records usually contain outcomes rather than descriptions of what happened. We present \method, which learns to describe item-specific behavior from these outcomes. A vision-language model generates a description, and a frozen language model scores its agreement with the clinical outcome, relevance to the item, abstention on unrelated video--item pairs, and clarity. Group relative policy optimization (GRPO) updates LoRA adapters using this semantic reward. On 13,379 videos covering 41 items, the method improves aggregate accuracy and the number of items with recall above 0.5. Errors remain in temporal direction, duration, and age-specific interpretations of behavior.
\end{abstract}


\section{Introduction}
\label{sec:intro}

\begin{figure*}[t]
\centering
\includegraphics[width=0.71\textwidth]{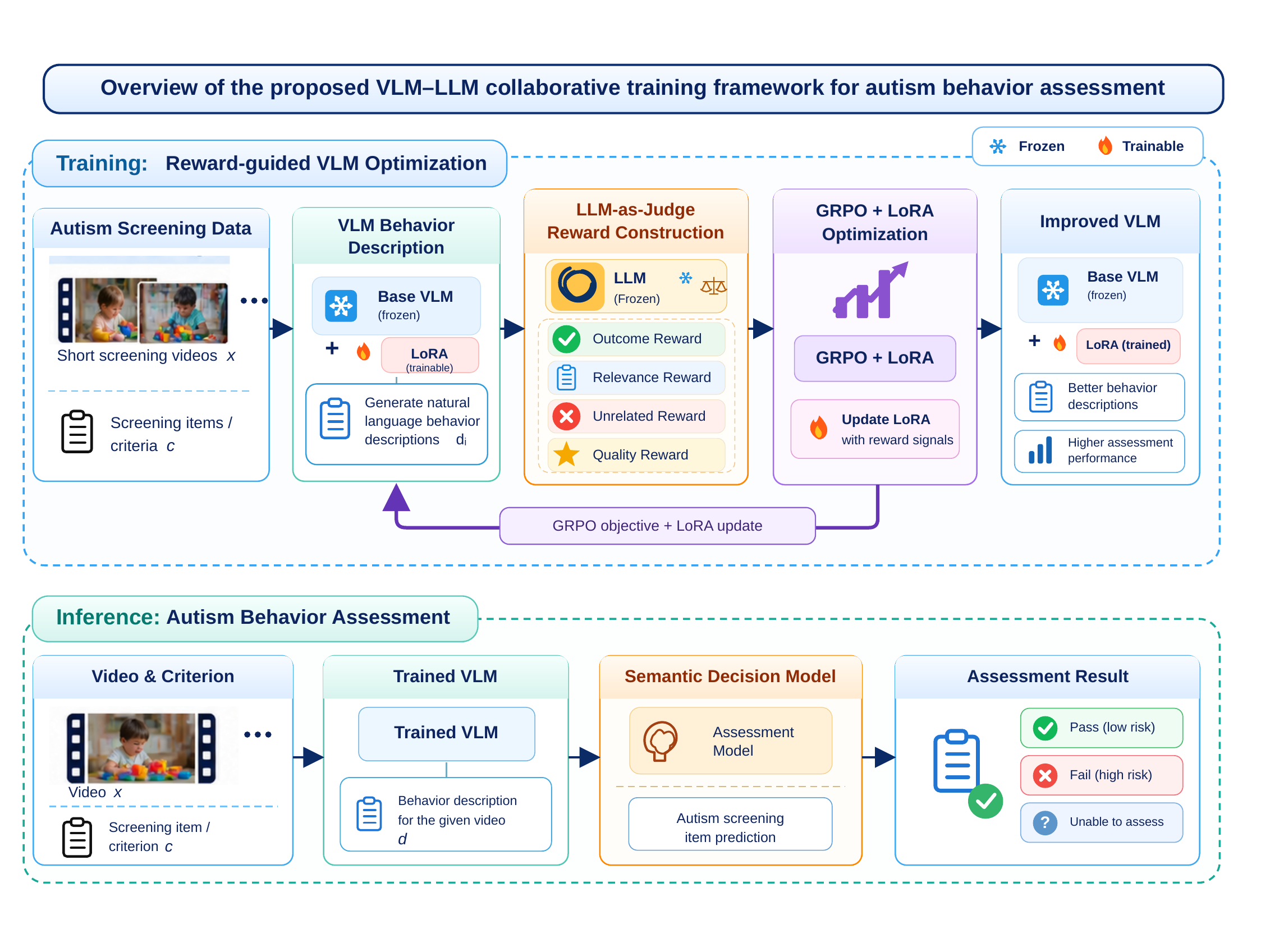}
\caption{\textbf{Overview of \method.} The VLM generates item-conditioned descriptions, a frozen LLM supplies semantic rewards, and GRPO updates rank-16 LoRA adapters; inference reuses the adapted descriptor before the final item decision.}
\label{fig:overview}
\vspace{-0.8em}
\end{figure*}

Naturalistic developmental videos capture gross-motor, fine-motor, language, adaptive, and social behaviors. Each clinical record includes an age-specific item, filming guidance, a pass criterion, a clinician-reviewed outcome, and an uploaded video. Most records lack a description of the behavior visible in the video. The outcome states the clinical decision but leaves its visual evidence implicit.

An item decision often depends on a small detail within a longer action. Walking backward requires the direction of movement, standing on one foot requires a sustained posture, and giving an object on request requires both the action and its social context. A broad account of the clip may describe the child accurately yet omit the evidence needed for the item. The item criterion must therefore guide both what the model describes and how the description is interpreted.

The same pass/fail outcome can arise from different visible behaviors. A failed item may reflect an absent or incomplete action, occlusion, or a behavior that was never elicited. A direct classifier $p(y\mid x,c)$ compresses these cases into one label. A single caption target is also unsuitable when several descriptions of a clip are valid. We instead generate an item-specific description before making the decision. When the video lacks the requested evidence, the model can respond ``unable to assess.''

Vision-language learning has progressed from transferable image-text representations such as CLIP and ALIGN \cite{clip,align} to generative multimodal models including Flamingo and BLIP-2 \cite{flamingo,blip2}. Instruction-tuned systems such as InstructBLIP, LLaVA, and MiniGPT-4 made free-form visual description a practical interface \cite{instructblip,llava,minigpt4}. VideoChat, Video-ChatGPT, and Video-LLaMA extended this interface to temporal inputs \cite{videochat,videochatgpt,videollama}, while LLaMA-VID, Video-LLaVA, and VideoLLaMA 2 improved long-context compression, alignment, and spatiotemporal modeling \cite{llamavid,videollava,videollama2}. General-purpose VLMs such as Qwen2-VL, InternLM-XComposer2.5, and LLaVA-OneVision further broaden resolution, context length, and task transfer \cite{qwen2vl,internlmx25,llavaonevision}.

Video understanding still depends on temporal evidence. MVBench measures video reasoning skills \cite{mvbench}. MovieChat and MA-LMM use memory to retain evidence \cite{moviechat,malmm}, VideoLLM-online processes streams incrementally \cite{videollmonline}, and video distillation improves video-language representations \cite{distillvideo}. TimeChat and TempCompass document difficulties with temporal order, direction, and duration \cite{timechat,tempcompass}. In biomedical applications, LLaVA-Med and Med-Flamingo adapt multimodal assistants to medical data \cite{llavamed,medflamingo}. BiomedCLIP learns biomedical representations \cite{biomedclip}, and CARE-VL explores autism-related screening \cite{carevl2025}. Video-R1 and R1-VL apply group-relative reinforcement learning to multimodal reasoning \cite{feng2025videor1,zhang2025r1vl}. We use clinician outcomes to train a description layer for developmental assessment.

We model assessment as two coupled semantic transformations: \emph{behavior description} and \emph{behavior decision}. A VLM first converts the video and item criterion into an inspectable description; a frozen LLM then interprets that description under the criterion. The same LLM also scores sampled descriptions for outcome consistency, item relevance, unrelated-pair abstention, and generation quality. These semantic rewards allow GRPO to adapt the VLM with rank-16 LoRA while leaving the visual backbone and judge frozen. Figure~\ref{fig:overview} summarizes the training and inference paths.

The description is the point at which visual recognition and clinical interpretation meet. It can state that an action occurred while also recording its direction, duration, or relation to an object. A clinician can then inspect whether the described behavior supports the item decision. During training, the outcome guides this intermediate description through the semantic reward. At inference, the decision model reads the description rather than the raw video, so errors in observation can be distinguished from errors in applying the criterion.

\noindent\textbf{Our contributions are:}

\noindent\textbf{1)} We formulate item-level developmental behavior assessment as an \emph{outcome-supervised description--decision} problem, introducing an inspectable natural-language evidence layer without requiring a unique expert-written caption for every video.

\noindent\textbf{2)} We construct a frozen-LLM semantic reward that combines outcome agreement, item relevance, unrelated-pair abstention, and generation quality, and optimize the descriptor with GRPO and rank-16 LoRA for parameter-efficient adaptation.

\noindent\textbf{3)} We evaluate the framework on 13,379 videos from 4,670 children, covering 41 developmental behaviors and two VLM generations. Accuracy and item-level recall coverage improve on both backbones. The remaining errors involve temporal direction, duration, and age-specific semantics.

\section{Method}
\label{sec:method}

\subsection{Problem formulation and two-stage inference}

Let $x$ denote a child video, $c$ an age-specific assessment item and its pass criterion, and $y\in\{\mathrm{pass},\mathrm{fail}\}$ the clinician-reviewed outcome. The VLM descriptor $\pi_\theta$ samples a behavior description
\begin{equation}
 d_i\sim\pi_\theta(\cdot\mid x,c),\qquad i=1,\ldots,G .
 \label{eq:descriptor}
\end{equation}
A frozen semantic decision model $h$ maps $(d_i,c)$ to pass, fail, or unable-to-assess. We use explicit item criteria in the judge prompt, following established LLM-as-a-Judge practice \cite{llmjudge}. Inference first produces a behavior description and then applies the semantic decision model, separating visual evidence from item interpretation.

For example, a criterion may ask whether a child pours an object out of a bottle. A useful description records the bottle, the child's manipulation of it, and whether an object visibly leaves the container. Details such as clothing or the bottle's decoration do not determine the outcome. The decision model compares the reported action with the criterion. If the relevant part of the action is outside the clip, it can return unable-to-assess instead of treating the missing observation as a failed attempt.

Several descriptions can accurately represent the same clip. A failed item also has no unique visual form: the target action may be absent, incomplete, replaced by another action, or not elicited. We optimize descriptions with outcome-level rewards. Each reward measures whether a description supports the clinician-reviewed outcome without prescribing its wording.

\subsection{Item-conditioned prompting and abstention}

The descriptor receives the exact item and pass criterion. For ``stand on one foot for 2 seconds,'' the prompt asks about the support leg and duration. For ``expresses no,'' it asks about visible refusal behavior. If the evidence is absent, occluded, or outside the recorded interval, the descriptor briefly states what is visible and says that the item cannot be assessed. Missing evidence therefore remains distinct from observed failure.

Item conditioning narrows the requested evidence without assuming that the child will express it in a single fixed way. A refusal, for instance, may appear as turning away or pushing an object. The description should name the visible movement before the decision model relates it to the age-specific criterion. This separation also makes an incorrect mapping visible: the model may observe the motion correctly but draw the wrong item conclusion.

Training also includes mismatched video--item pairs $(x,c')$, where $c'$ names a behavior unrelated to the video. The descriptor earns a reward for abstaining on these pairs. This teaches it to distinguish missing evidence from evidence of failure.

\subsection{LLM-as-a-Judge semantic reward}

For each sampled description $d_i$, a frozen Qwen3-8B judge evaluates four binary properties. \textbf{Outcome reward} checks both evidence sufficiency and whether the item decision implied by $d_i$ agrees with the clinician outcome $y$. A candidate receives $r_i^{\rm out}=1$ only when both conditions hold. \textbf{Unrelated-pair reward} is active on mismatched pairs and assigns $r_i^{\rm unrel}=1$ only when the descriptor correctly abstains. \textbf{Relevance reward} $r_i^{\rm rel}$ checks whether the text actually addresses the requested item, while \textbf{quality reward} $r_i^{\rm qual}$ rejects garbled or internally contradictory generations. The total semantic reward is
\begin{equation}
 r_i=r_i^{\rm out}+r_i^{\rm unrel}+r_i^{\rm rel}+r_i^{\rm qual}.
 \label{eq:reward}
\end{equation}
Outcome agreement alone produced irrelevant or malformed responses, particularly under the strong pass-label imbalance. Matched records receive an outcome reward, and mismatched pairs receive an abstention reward. Relevance and quality are scored for both. The four binary terms allow varied wording while keeping descriptions tied to the item. The judge remains frozen throughout training.

\subsection{GRPO optimization}

For every input, the descriptor samples $G=16$ candidate descriptions. Their rewards are standardized within the group,
\begin{equation}
 \hat A_i=\frac{r_i-\mu_r}{\sigma_r+\delta},
 \label{eq:adv}
\end{equation}
where $\mu_r$ and $\sigma_r$ are the group reward mean and standard deviation. The clipped objective is
\begin{equation}
\mathcal{J}(\theta)=\frac{1}{G}\sum_i\min\!\left(\rho_i\hat A_i,
\mathrm{clip}(\rho_i,1-\epsilon,1+\epsilon)\hat A_i\right)-\beta D_{\rm KL},
\label{eq:grpo}
\end{equation}
Here $\rho_i$ is the current-to-old policy ratio, and the KL term anchors the reference policy. GRPO uses the sampled group as its baseline without a separate critic \cite{shao2024deepseekmath}. Similar rewards across all candidates give the update little directional signal. We observe this pattern in repeated temporal-direction errors (Sec.~\ref{sec:exp}).

\subsection{Parameter-efficient adaptation}

Only the language-model component of the VLM is adapted with LoRA \cite{hu2021lora}. For a frozen weight matrix $W$, LoRA represents the update as
\begin{equation}
 W'=W+BA,\qquad \mathrm{rank}(BA)=r\ll\min(d_{\rm in},d_{\rm out}),
 \label{eq:lora}
\end{equation}
with rank $r=16$. The visual backbone, pretrained VLM weights, and judge remain frozen. Training uses AdamW ($10^{-5}$), batch size 8, two epochs, five iterations, and LLM-diversified item prompts. Preliminary full-parameter tuning was unstable on these data. Rank-16 LoRA trained stably and is used in all reported models.

\begin{table}[t]
\centering
\small
\caption{Training configuration.}
\label{tab:train}
\setlength{\tabcolsep}{5pt}
\begin{tabular}{lc}
\toprule
Setting & Value \\
\midrule
Group size $G$ & 16 \\
LoRA target & VLM language layers \\
LoRA rank & 16 \\
Optimizer / learning rate & AdamW / $10^{-5}$ \\
Batch size / epochs & 8 / 2 \\
Training iterations & 5 \\
Train/test protocol & per-item 90/10 \\
\bottomrule
\end{tabular}
\end{table}

\section{Experiments}
\label{sec:exp}

\subsection{Dataset and evaluation protocol}

The dataset contains 13,379 naturalistic developmental videos from 4,670 children, collected at child-health and developmental clinics after guardian consent. Ages range from 2 to 42 months, with a mean of 12.73 months. The cohort includes 2,313 boys and 2,357 girls. We evaluate 41 items at ages 3, 6, 8, 10, 12, 18, 24, and 30 months. Each example includes the item, filming guidance, pass criterion, clinician-reviewed outcome, and a video uploaded by a caregiver or clinician. The videos have no dense behavior descriptions.

The 41 items cover gross-motor, fine-motor, social, and adaptive behaviors. They include rolling from supine to prone, walking backward, standing on one foot, transferring blocks, giving an object on request, and expressing refusal. A shared description layer can represent the different actions and criteria across items.

Class prevalence is markedly imbalanced. For the six-month ``reach for an object'' item, 369 of 370 videos have a pass label. The same motion can also carry different meanings across ages and item criteria. We report item-wise precision and recall alongside aggregate accuracy. Each item uses 90\% of its videos for training and 10\% for testing. All prompt and adaptation comparisons use this split protocol.

\begin{table}[t]
\centering
\small
\caption{Dataset summary.}
\label{tab:data}
\setlength{\tabcolsep}{5pt}
\begin{tabular}{lc}
\toprule
Statistic & Value \\
\midrule
Children & 4,670 \\
Videos & 13,379 \\
Age range / mean & 2--42 / 12.73 months \\
Male / female & 2,313 / 2,357 \\
Key developmental ages & 8 \\
Evaluated behavior items & 41 \\
Dense behavior captions & Not available \\
\bottomrule
\end{tabular}
\end{table}

\subsection{Evaluation protocol}

We evaluate descriptions with a frozen LLM-as-a-Judge. To assess relevance, it reads the description and item prompt and returns a binary decision. To assess behavior accuracy, it first compares the described evidence with the criterion, then returns a constrained yes/no decision. An unrelated video--item pair is correct when the system responds ``unable to assess.'' Item-wise recall and precision measure evidence recovery; aggregate accuracy measures agreement with clinician-reviewed outcomes.

The evaluation uses the same item text and pass criterion across the base and adapted models. The judge's first turn considers whether the description contains enough behavior evidence to apply the criterion. Its second turn provides a constrained answer that can be parsed consistently. This protocol measures whether the generated text supports an item decision; it does not require the text to match a particular sentence. The unrelated-pair test separately checks whether the model avoids asserting a behavior that the video does not show.

\begin{figure*}[t]
\centering
\begin{minipage}[t]{0.47\textwidth}
\centering
\includegraphics[width=\linewidth]{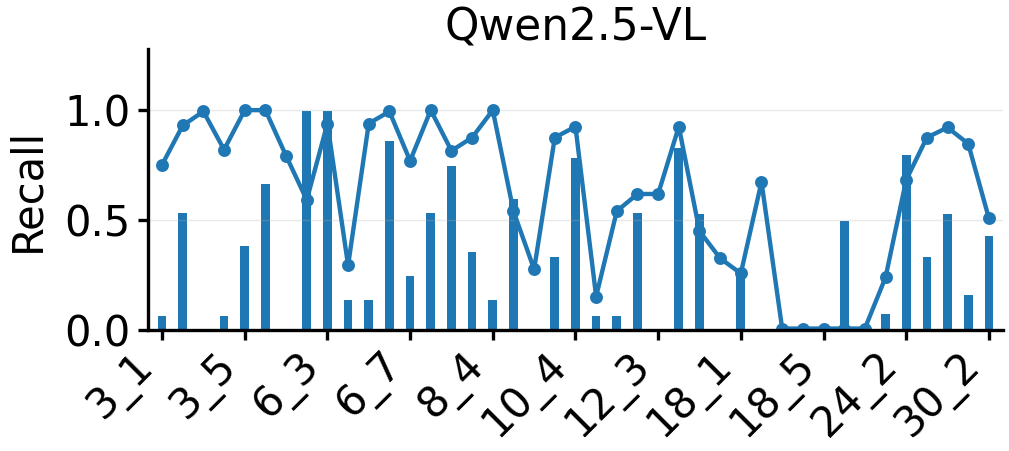}\par\vspace{0.25em}
{\fontsize{9}{10}\selectfont (a) Qwen2.5-VL}
\end{minipage}\hfill
\begin{minipage}[t]{0.47\textwidth}
\centering
\includegraphics[width=\linewidth]{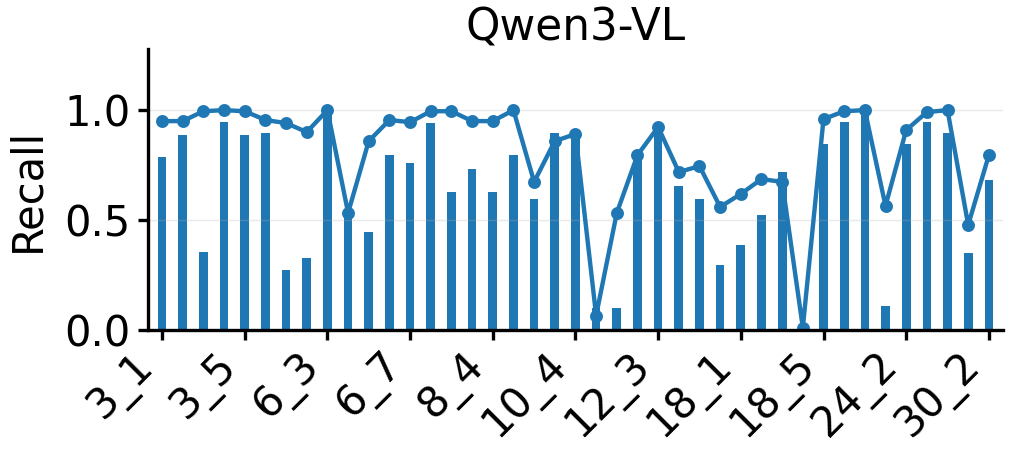}\par\vspace{0.25em}
{\fontsize{9}{10}\selectfont (b) Qwen3-VL}
\end{minipage}
\caption{\textbf{Item-level recall before and after adaptation.} Bars denote the base VLM and lines the GRPO-LoRA model over 41 age--item codes; adapted Qwen3-VL reaches recall $>0.5$ on 38/41 items.}
\label{fig:itemwise}
\vspace{-0.35em}
\end{figure*}

\subsection{Prompt specificity before adaptation}

We first test general-purpose VLMs with broad prompts. The study includes InternLM-XComposer2.5 \cite{internlmx25}, Qwen2.5-VL-Instruct-7B \cite{qwen25vl}, and Qwen3-VL-Instruct-8B. Each prompt requests a summary of one developmental domain: gross motor, fine motor, language, personal-social, or adaptive behavior. The resulting descriptions often include fluent but irrelevant observations and omit details needed for an individual item. For Qwen3-VL, 5 of 41 items reach recall above 0.5: 12\_4 (0.513), 3\_1 (0.520), 18\_7 (0.570), 24\_4 (0.667), and 3\_6 (0.713).

The item-specific prompt states the exact item and pass criterion and asks the model to abstain when evidence is absent. With this prompt, 30 of 41 items reach recall above 0.5. Seven exceed 0.9: 8\_1 (0.944), 12\_3 (0.944), 3\_4 (0.950), 18\_6 (0.950), 24\_3 (0.950), 6\_3 (1.000), and 18\_7 (1.000). The descriptions are also shorter and more relevant.

The assessment items differ in the detail needed for a decision. Sitting independently centers on posture, turning pages requires an ordered hand action, and drinking from a cup requires object use. Broad domain prompts can mention these scenes without resolving the required detail. Item-specific prompts make the relevant action explicit, which explains the large coverage increase before any parameter update. The remaining low-recall items show where prompt specificity alone does not recover sufficient visual or temporal evidence.

\subsection{GRPO-LoRA adaptation}

Table~\ref{tab:accuracy} shows gains on both backbones. Qwen2.5-VL accuracy rises from 0.669 to 0.817 (+22.12\% relative), and Qwen3-VL rises from 0.808 to 0.890 (+10.15\% relative). The adapted Qwen2.5-VL approaches the unadapted Qwen3-VL. Adapted Qwen3-VL reaches the highest accuracy.

\begin{table}[t]
\centering
\small
\caption{Aggregate accuracy under the per-item 90/10 split.}
\label{tab:accuracy}
\setlength{\tabcolsep}{4.5pt}
\begin{tabular}{lccc}
\toprule
Backbone & Base & +GRPO-LoRA & Abs. gain \\
\midrule
Qwen2.5-VL & 0.669 & \textbf{0.817} & +0.148 \\
Qwen3-VL & 0.808 & \textbf{0.890} & +0.082 \\
\bottomrule
\end{tabular}
\vspace{-0.4em}
\end{table}

Prompting and adaptation each improve item coverage. The number of items with recall above 0.5 rises from 5/41 with broad prompts to 30/41 with item-specific prompts, then to 38/41 with \method (Table~\ref{tab:coverage}).

\begin{table}[t]
\centering
\small
\caption{Behavior coverage counted by items with recall $>0.5$.}
\label{tab:coverage}
\setlength{\tabcolsep}{3.5pt}
\begin{tabular}{lc}
\toprule
Setting & Items with recall $>0.5$ \\
\midrule
Generic domain prompt & 5 / 41 \\
Item-specific prompt & 30 / 41 \\
Item-specific prompt + \method & \textbf{38 / 41} \\
\bottomrule
\end{tabular}
\vspace{-0.5em}
\end{table}

Item-conditioned prompts produced shorter and more relevant descriptions than broad prompts. With outcome reward alone, mismatched pairs often produced irrelevant or malformed responses. Full-parameter tuning was also less stable than rank-16 LoRA. The final design combines item-specific prompting, four semantic reward terms, and LoRA adaptation.

\balance
\subsection{Item-level behavior and failure modes}

Figure~\ref{fig:itemwise} shows item-level gains for both backbones. Qwen2.5-VL starts with lower coverage and has the larger absolute accuracy gain. Qwen3-VL starts higher and reaches the best final accuracy and widest coverage (Table~\ref{tab:accuracy}).

Three items remain below 0.5 recall: walking backward (18\_4, 0.000), expressing ``no'' (10\_5, 0.053), and standing on one foot for 2 s (30\_1, 0.471). Walking backward and standing on one foot require temporal grounding. The model may recognize walking but confuse its direction, or miss how long a posture lasts. When all 16 sampled descriptions repeat an error, GRPO has little signal to favor a correction. Frame reversal, order perturbation, and duration examples may help address this pattern. Expressing refusal requires a different interpretation: a young child may turn away or push an object instead of shaking their head.

The descriptions reveal three error types. The model may miss a temporal attribute, misinterpret a visible action under the item criterion, or lack the requested evidence. These cases call for temporal examples, criterion-specific examples, and abstention training, respectively.

\subsection{Discussion and stratified analysis}

Item conditioning directs the model toward the action, direction, duration, object interaction, or response specified by the criterion. GRPO-LoRA improves how consistently it describes that evidence. Coverage rises from 5/41 items with broad prompts to 30/41 with item-specific prompts and 38/41 after adaptation. Qwen3-VL improves despite its stronger starting point.

The description layer makes visual-grounding and semantic-decision errors easier to distinguish. It also retains ``unable to assess'' for missing evidence. Each error type points to a specific training need: temporally difficult clips, criterion-specific examples, or mismatched pairs.

The abstention test addresses a separate failure. Most clinical outcomes in this collection are passes, so a model can appear accurate by predicting that a requested behavior occurred. Pairing a video with an unrelated item tests whether the description is grounded in the clip. A correct response briefly reports the visible content and withholds the item decision.

Sex-stratified precision is broadly similar. For items 10\_5, 18\_4, and 18\_7, the test subsets have no valid female examples, so those gaps cannot be interpreted as performance differences. The per-item split measures performance within the collection distribution. Child-disjoint and cross-institution splits would test transfer to new children and recording settings.

\section{Conclusion}
\label{sec:conclusion}

Routine developmental records provide clinician outcomes but rarely describe the behavior visible in each video. \method addresses this gap by separating behavior description from item decision. Frozen-LLM rewards guide GRPO-LoRA adaptation without caption labels. On 13,379 videos covering 41 items, item-specific prompts raise the number of items with recall above 0.5 from 5 to 30, and adaptation raises it to 38. Accuracy also improves on both VLM backbones. The descriptions make decisions easier to inspect and expose remaining errors in temporal direction, duration, and age-specific interpretation.


\clearpage
\begin{thebibliography}{40}

\bibitem{clip}
A. Radford, J. W. Kim, C. Hallacy, A. Ramesh, G. Goh, S. Agarwal, G. Sastry, A. Askell, P. Mishkin, J. Clark, G. Krueger, and I. Sutskever, ``Learning transferable visual models from natural language supervision,'' in \emph{Proc. ICML}, 2021, pp. 8748--8763.

\bibitem{align}
C. Jia, Y. Yang, Y. Xia, Y.-T. Chen, Z. Parekh, H. Pham, Q. Le, Y.-H. Sung, Z. Li, and T. Duerig, ``Scaling up visual and vision-language representation learning with noisy text supervision,'' in \emph{Proc. ICML}, 2021, pp. 4904--4916.

\bibitem{flamingo}
J.-B. Alayrac, J. Donahue, P. Luc, \emph{et al.}, ``Flamingo: A visual language model for few-shot learning,'' in \emph{Advances in Neural Information Processing Systems}, vol. 35, 2022.

\bibitem{blip2}
J. Li, D. Li, S. Savarese, and S. Hoi, ``BLIP-2: Bootstrapping language-image pre-training with frozen image encoders and large language models,'' in \emph{Proc. ICML}, 2023, pp. 19730--19742.

\bibitem{instructblip}
W. Dai, J. Li, D. Li, A. M. H. Tiong, J. Zhao, W. Wang, B. Li, P. Fung, and S. Hoi, ``InstructBLIP: Towards general-purpose vision-language models with instruction tuning,'' in \emph{Advances in Neural Information Processing Systems}, vol. 36, 2023.

\bibitem{llava}
H. Liu, C. Li, Q. Wu, and Y. J. Lee, ``Visual instruction tuning,'' in \emph{Advances in Neural Information Processing Systems}, vol. 36, 2023.

\bibitem{minigpt4}
D. Zhu, J. Chen, X. Shen, X. Li, and M. Elhoseiny, ``MiniGPT-4: Enhancing vision-language understanding with advanced large language models,'' arXiv:2304.10592, 2023.

\bibitem{videochat}
K. Li, Y. He, Y. Wang, Y. Li, W. Wang, P. Luo, Y. Wang, L. Wang, and Y. Qiao, ``VideoChat: Chat-centric video understanding,'' arXiv:2305.06355, 2023.

\bibitem{videochatgpt}
M. Maaz, H. Rasheed, S. Khan, and F. S. Khan, ``Video-ChatGPT: Towards detailed video understanding via large vision and language models,'' arXiv:2306.05424, 2023.

\bibitem{videollama}
H. Zhang, X. Li, and L. Bing, ``Video-LLaMA: An instruction-tuned audio-visual language model for video understanding,'' arXiv:2306.02858, 2023.

\bibitem{llamavid}
Y. Li, C. Wang, and J. Jia, ``LLaMA-VID: An image is worth 2 tokens in large language models,'' arXiv:2311.17043, 2023.

\bibitem{videollava}
B. Lin, Y. Ye, B. Zhu, J. Cui, M. Ning, P. Jin, and L. Yuan, ``Video-LLaVA: Learning united visual representation by alignment before projection,'' arXiv:2311.10122, 2023.

\bibitem{videollama2}
Z. Cheng \emph{et al.}, ``VideoLLaMA 2: Advancing spatial-temporal modeling and audio understanding in video-LLMs,'' arXiv:2406.07476, 2024.

\bibitem{qwen2vl}
P. Wang \emph{et al.}, ``Qwen2-VL: Enhancing vision-language model's perception of the world at any resolution,'' arXiv:2409.12191, 2024.

\bibitem{internlmx25}
P. Zhang \emph{et al.}, ``InternLM-XComposer-2.5: A versatile large vision language model supporting long-contextual input and output,'' arXiv:2407.03320, 2024.

\bibitem{llavaonevision}
B. Li \emph{et al.}, ``LLaVA-OneVision: Easy visual task transfer,'' arXiv:2408.03326, 2024.

\bibitem{mvbench}
K. Li \emph{et al.}, ``MVBench: A comprehensive multi-modal video understanding benchmark,'' in \emph{Proc. CVPR}, 2024, pp. 22195--22206.

\bibitem{moviechat}
E. Song \emph{et al.}, ``MovieChat: From dense token to sparse memory for long video understanding,'' in \emph{Proc. CVPR}, 2024, pp. 18221--18232.

\bibitem{malmm}
B. He \emph{et al.}, ``MA-LMM: Memory-augmented large multimodal model for long-term video understanding,'' in \emph{Proc. CVPR}, 2024, pp. 13504--13514.

\bibitem{videollmonline}
J. Chen \emph{et al.}, ``VideoLLM-online: Online video large language model for streaming video,'' in \emph{Proc. CVPR}, 2024, pp. 18407--18418.

\bibitem{distillvideo}
Y. Zhao \emph{et al.}, ``Distilling vision-language models on millions of videos,'' in \emph{Proc. CVPR}, 2024, pp. 13106--13116.

\bibitem{timechat}
S. Ren, L. Yao, S. Li, X. Sun, and L. Hou, ``TimeChat: A time-sensitive multimodal large language model for long video understanding,'' in \emph{Proc. CVPR}, 2024.

\bibitem{tempcompass}
Y. Liu \emph{et al.}, ``TempCompass: Do video LLMs really understand videos?'' arXiv:2403.00476, 2024.

\bibitem{llavamed}
C. Li \emph{et al.}, ``LLaVA-Med: Training a large language-and-vision assistant for biomedicine in one day,'' arXiv:2306.00890, 2023.

\bibitem{medflamingo}
M. Moor \emph{et al.}, ``Med-Flamingo: A multimodal medical few-shot learner,'' arXiv:2307.15189, 2023.

\bibitem{biomedclip}
S. Zhang \emph{et al.}, ``BiomedCLIP: A multimodal biomedical foundation model pretrained from fifteen million scientific image-text pairs,'' arXiv:2303.00915, 2023.

\bibitem{carevl2025}
C.-H. Yoo, J.-H. Yoo, and J. Jang, ``CARE-VL: A domain-specialized vision-language model for early ASD screening,'' in \emph{MICCAI}, 2025.

\bibitem{feng2025videor1}
K. Feng \emph{et al.}, ``Video-R1: Reinforcing video reasoning in MLLMs,'' in \emph{Advances in Neural Information Processing Systems}, 2025.

\bibitem{zhang2025r1vl}
J. Zhang \emph{et al.}, ``R1-VL: Learning to reason with multimodal large language models via step-wise group relative policy optimization,'' in \emph{Proc. ICCV}, 2025.

\bibitem{llmjudge}
L. Zheng \emph{et al.}, ``Judging LLM-as-a-Judge with MT-Bench and Chatbot Arena,'' arXiv:2306.05685, 2023.

\bibitem{shao2024deepseekmath}
Z. Shao \emph{et al.}, ``DeepSeekMath: Pushing the limits of mathematical reasoning in open language models,'' arXiv:2402.03300, 2024.

\bibitem{hu2021lora}
E. J. Hu \emph{et al.}, ``LoRA: Low-rank adaptation of large language models,'' in \emph{ICLR}, 2022.

\bibitem{qwen25vl}
Qwen Team, ``Qwen2.5-VL Technical Report,'' arXiv:2502.13923, 2025.
\end{thebibliography}
\end{document}